\documentclass[sigconf, nonacm]{acmart}
\AtBeginDocument{%
  }

\usepackage{hyperref}       
\usepackage{url}            
\usepackage{booktabs}       
\usepackage{amsfonts}       
\usepackage{nicefrac}       
\usepackage{microtype}      
\usepackage{xcolor}         
\usepackage{graphicx} 
\usepackage{subcaption}
\usepackage{bbm}
\usepackage{enumitem} 
\usepackage{multirow}

\usepackage{atbegshi}
\usepackage{lineno}
\begin{document}

\title{THOR: Inductive Link Prediction over Hyper-Relational Knowledge Graphs}


\author{Weijian Yu}
\affiliation{%
  \institution{University of Macau}
  \city{Macau}
  \country{China}}
\email{yc47946@um.edu.mo}

\author{Yuhuan Lu}
\affiliation{%
  \institution{Khalifa University}
  \city{Abu Dhab}
  \country{UAE}
}
\email{yuhuan.lu@ku.ac.ae}

\author{Dingqi Yang}
\authornote{Corresponding author}
\affiliation{%
 \institution{University of Macau}
  \city{Macau}
  \country{China}}
  \email{dingqiyang@um.edu.mo}





\renewcommand{\shortauthors}{}


\begin{abstract}
Knowledge graphs (KGs) have become a key ingredient supporting a variety of applications. Beyond the traditional triplet representation of facts where a relation connects two entities, modern KGs observe an increasing number of hyper-relational facts, where an arbitrary number of qualifiers associated with a triplet provide auxiliary information to further describe the rich semantics of the triplet, which can effectively boost the reasoning performance in link prediction tasks. However, existing link prediction techniques over such hyper-relational KGs (HKGs) mostly focus on a transductive setting, where KG embedding models are learned from the specific vocabulary of a given KG and subsequently can only make predictions within the same vocabulary, limiting their generalizability to previously unseen vocabularies. Against this background, we propose THOR, an induc\underline{T}ive link prediction technique for \underline{H}yper-relational kn\underline{O}wledge g\underline{R}aphs. Specifically, we first introduce both relation and entity foundation graphs, modeling their fundamental inter- and intra-fact interactions in HKGs, which are agnostic to any specific relations and entities. Afterward, THOR is designed to learn from the two foundation graphs with two parallel graph encoders followed by a transformer decoder, which supports efficient masked training and fully-inductive inference. We conduct a thorough evaluation of THOR in hyper-relational link prediction tasks on 12 datasets with different settings. Results show that THOR outperforms a sizable collection of baselines, yielding 66.1\%, 55.9\%, and 20.4\% improvement over the best-performing rule-based, semi-inductive, and fully-inductive techniques, respectively. A series of ablation studies also reveals our key design factors capturing the structural invariance transferable across HKGs for inductive tasks.
\end{abstract}





\maketitle

\section{Introduction}
Knowledge graphs (KGs) have been widely used to support various real-world applications ranging from question answering to recommendation systems \cite{wang2017knowledge, ji2021survey}. In a typical KG, factual knowledge is traditionally represented as a set of triplets and each triplet $\left ( h, r, t \right )$ consists of a head entity $h$, a tail entity $t$, and a relation $r$ connecting them, such as \textit{(Albert Einstein, educated at, ETH Zurich)}. To efficiently reason over such KGs, link prediction tasks are widely adopted to predict a missing entity in a query fact $\left ( h, r, \left [ MASK \right ]  \right ) $, where the $[MASK]$ token represents the missing entity. However, it has been widely revealed that the traditional triplet-based KGs oversimplify the real-world facts, which are often more complex beyond the triplet structure \cite{rosso2020beyond}. Specifically, modern KGs contain an increasing number of hyper-relational facts, where a primary triplet $(h, r, t)$ is associated with several qualifiers, represented as key-value (relation-entity) pairs $(k,v)$, providing auxiliary information to further describe the rich semantics of the primary triplet. For example, \textit{[(Albert Einstein, educated at, ETH Zurich), {(academic degree, Bachelor of Science), (academic major, mathematics education)}]} further indicates the academic degree and academic major beyond the aforementioned triple fact. Such qualifiers are very useful in boosting link prediction performance over KGs, by providing additional clues for predicting the missing entity \cite{rosso2020beyond}.

Despite the recent advancement in link prediction over hyper-relational KGs (HKG) \cite{galkin2020message, wang2021link, rosso2020beyond, luo2023hahe, guan2019link,guan2020neuinfer}, existing techniques mostly consider a transductive setting, where models learn entity and relation embeddings from the specific vocabulary of a given KG to capture the structural and semantic properties, and subsequently can only make prediction within the same vocabulary, failing to generalize to unseen entities and relations. However, the dynamics of modern KGs imply continuously emerging entities and relations over time, while these transductive link prediction techniques thus show limited generalizability by failing to generalize to previously unseen vocabularies. In this context, a few existing works design inductive link prediction techniques under either a \textit{semi-inductive setting} with a consistent relation vocabulary but arbitrary (even disjoint) entity vocabularies \cite{yin2025inductive}, or a more generalized \textit{fully-inductive setting} with both arbitrary (even disjoint) relation and entity vocabularies \cite{galkin2023towards}. Different from the transductive link prediction techniques that rely on specific embeddings of entities and relations in a KG, the inductive techniques need to learn the structural invariance transferable across KGs and can thus generalize to arbitrary entity and relation vocabularies.

In the current literature, existing inductive link prediction techniques face the following key challenges when handling HKGs. First, these techniques mainly consider triple facts by designing a relation graph modeling the fundamental interactions (edges) between relations, such as head-to-tail or tail-to-head interactions \cite{geng2023relational, galkin2023towards} or membership of certain graph motifs \cite{huang2025expressive}. As the edge of this relation graph is agnostic about any specific relations, models learnt from this graph thus support inductive inference. However, this relation graph overlooks the sophistication of the HKGs with qualifiers, which significantly enrich yet also complicate the interactions between entities and relations beyond the triple facts, as the qualifiers associated with the primary triplet are also represented as relation-entity pairs. Second, existing approaches mostly design graph encoders to learn structural invariance from the relation graph by contrasting a training triple fact against randomly corrupted negative samples (randomly corrupting one of the two entities in the triple fact) \cite{zhu2021neural, lee2023ingram, galkin2023towards, galkin2024foundation, geng2023relational, zhang2025trix}. However, lessons learnt from transductive techniques \cite{yu2024robust} revealed that such a random corruption negative sampling scheme is significantly inefficient for HKGs due to the large corruption space, where the corruption space increases exponentially with the number of elements in a hyper-relational fact.

To address the above challenges, we propose THOR, a fully-induc\underline{T}ive link prediction technique for \underline{H}yper-relational kn\underline{O}wledge g\underline{R}aphs. First, we design two foundation graphs capturing both fundamental interactions between relations and entities in an HKG, respectively. Since both foundation graphs model the position-wise interactions of elements in the HKG, they are agnostic to any specific relations and entities, thus supporting fully-inductive inference. Second, for inductive learning from the relation and entity foundation graphs, we first design two parallel graph encoders based on Neural Bellman-Ford Networks (NBFNet) \cite{zhu2021neural}, which is a path-based graph neural architecture for generic inductive link prediction, followed by a transformer decoder, fully supporting masked training without the need for the random corruption negative sampling process. Our contributions are as follows:

\begin{itemize}[leftmargin=*]
    \item We introduce both relation and entity foundation graphs for HKGs, which model their fundamental inter- and intra-fact interactions and are agnostic to any specific relations and entities, thus being capable of supporting fully-inductive inference. 

    \item We propose an inductive link prediction technique THOR, learning from the two foundation graphs with two parallel NBFNet-based graph encoders, followed by a transformer decoder, fully supporting efficient masked training without the need for the expensive random corruption negative sampling process.

    \item We conduct a thorough evaluation of THOR in different settings on 12 datasets from different domains. Results show that THOR outperforms a sizable collection of 16 baselines in different link prediction tasks, yielding 66.1\%, 55.9\%, and 20.4\% improvement over the best-performing rule-based, semi-inductive, and fully-inductive techniques, respectively. In particular, in cross-domain fully-inductive tasks, it outperforms the best baselines by 32.5\%. 
    
    \item A series of ablation studies further verify our design choices of THOR, revealing key design factors capturing the structural invariance transferable across HKG: 1) qualifiers are indeed useful for inductive settings, but defining their fundamental interactions is tricky; we found that using relation-key interactions and exhaustive entity interactions is most effective; 2) avoiding negative samples is also very useful, yielding 57.9\% improvement.
\end{itemize}


\section{Preliminary}
\label{sec:preliminary}
We first define our key terminology, followed by problem settings.

\noindent \textbf{Hyper-relational Knowledge Graph (HKG).} Given an entity set $E$ and a relation set $R$, a hyper-relational KG $\mathcal{G}=\left( E, R, F \right)$ consists of a set of fact $F = \left \{ \left [ \left ( h, r, t  \right ), \left \{ \left ( k_{i}, v_{i} \right )   \right \}_{i=1}^{n} \right ] \mid h, t, v_i \in E, r, k_i \in R   \right \} $, where $\left ( h, r, t  \right )$ refers to the primary triplet of this fact, and $\left \{ \left ( k_{i}, v_{i} \right )   \right \}_{i=1}^{n}$ refer to $n$ key-value pairs providing additional information about the primary triplet.

\noindent \textbf{Link prediction over HKGs.} Given a query hyper-relational fact with a missing entity represented by a $\left [ \mathit{MASK}  \right ]$ token, the link prediction task aims at predicting the masked entity. In our setting, the masked entity can be at any position in a fact, i.e., $h$, $t$, or $v$.

\noindent \textbf{Transductive link prediction task.} In the transductive setting, models learn from a training KG $\mathcal{G}_{train}=( E_{train}, R_{train}, F_{train} )$, and make predictions on query facts from an inference KG $\mathcal{G}_{inf}=( E_{inf}, R_{inf}, F_{inf} )$, where the inference KG are defined on the same entity and relation vocabularies as the training KG, i.e., $ E_{inf} \subseteq E_{train} $ and $R_{inf} \subseteq R_{train}$.

\noindent \textbf{Inductive link prediction task.} In the inductive setting, the training KG and the inference KG do not share the same entity and relation vocabularies. According to their shared elements, there are mainly two settings of inductive link production tasks\footnote{We note that a few works adopt different definitions of inductive settings; please see Appendix \ref{app_diff_inductive} for details.}.


\begin{itemize}[leftmargin=*]
    \item \textbf{\textit{Semi-inductive link prediction task.}} The semi-inductive setting assumes the same relation vocabulary for the training and inference KGs $R_{train} = R_{inf}$, but different entity vocabularies between the training and inference KGs. The entity vocabulary in the inference graph contains unseen entities in the training stage, where the inference entity set could be an extension of the training entity set $E_{train} \subset E_{inf}$, or they can be completely disjoint $E_{train} \cap E_{inf} = \emptyset $.
    
    \item \textbf{\textit{Fully-inductive link prediction task.}} The fully-inductive setting assumes both different relation and entity vocabularies between the training and inference KGs, where both relation and entity vocabularies in the inference graph contain unseen relations and entities in the training stage, respectively. In the hardest setting, both relation and entity vocabularies are disjoint between the training and inference KGs, i.e., $R_{train} \cap R_{inf} = \emptyset $ and $E_{train} \cap E_{inf} = \emptyset $.

\end{itemize}

In this work, we focus on the (hardest) fully-inductive case with both disjoint relation and entity vocabularies between the training and inference KGs. \textit{Note that since the harder inductive setting is strictly a superset of the easier inductive (or transductive) settings, fully-inductive link prediction techniques are thus by design applicable in all other settings.}

\begin{figure*}
  \centering
  \includegraphics[width=0.95\linewidth]{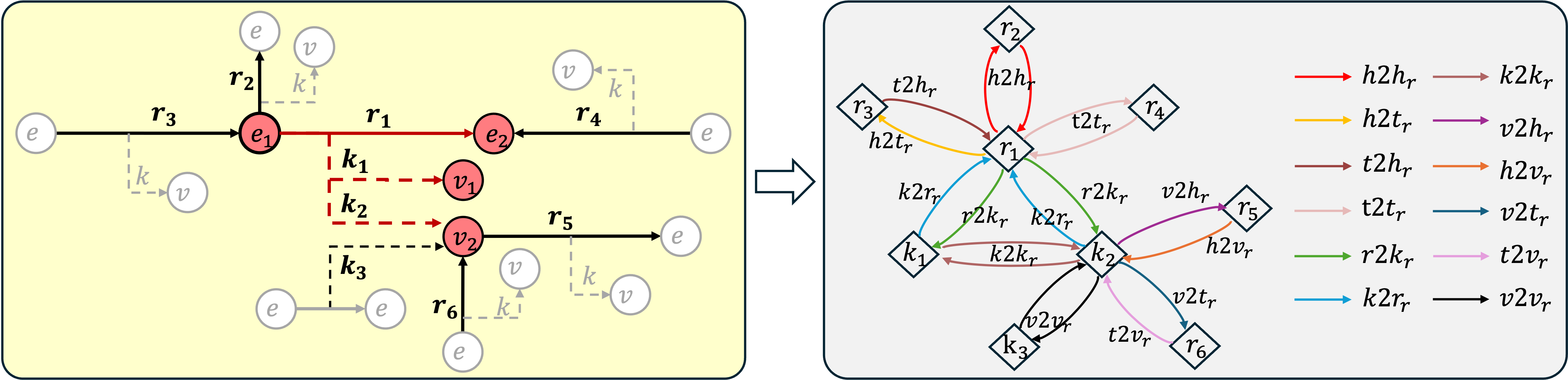}
  \vspace{-0.5em}
  \caption{Relation foundation graph with different fundamental interactions. The plot focuses on the relations of the fact highlighted in red, illustrating all the fundamental interactions involving its relations $r_1$, $k_1$, and $k_2$ only. Interactions among other relations also exist (e.g., $h2t_r$ and $t2h_r$ interactions between $r_2$ and $r_3$), but are not shown for the clarity of visualization.}
  \vspace{-0.5em}
  \label{foundation_relation}
\end{figure*}

\section{Method}
\label{method}
In this section, we introduce THOR, a fully-induc\underline{T}ive link prediction technique for \underline{H}yper-relational kn\underline{O}wledge g\underline{R}aphs. In the following, we first introduce the relation and entity foundation graphs, and then present THOR learning from both foundation graphs for fully-inductive link prediction over HKGs.

\begin{figure}
    \centering
    \includegraphics[width=\linewidth]{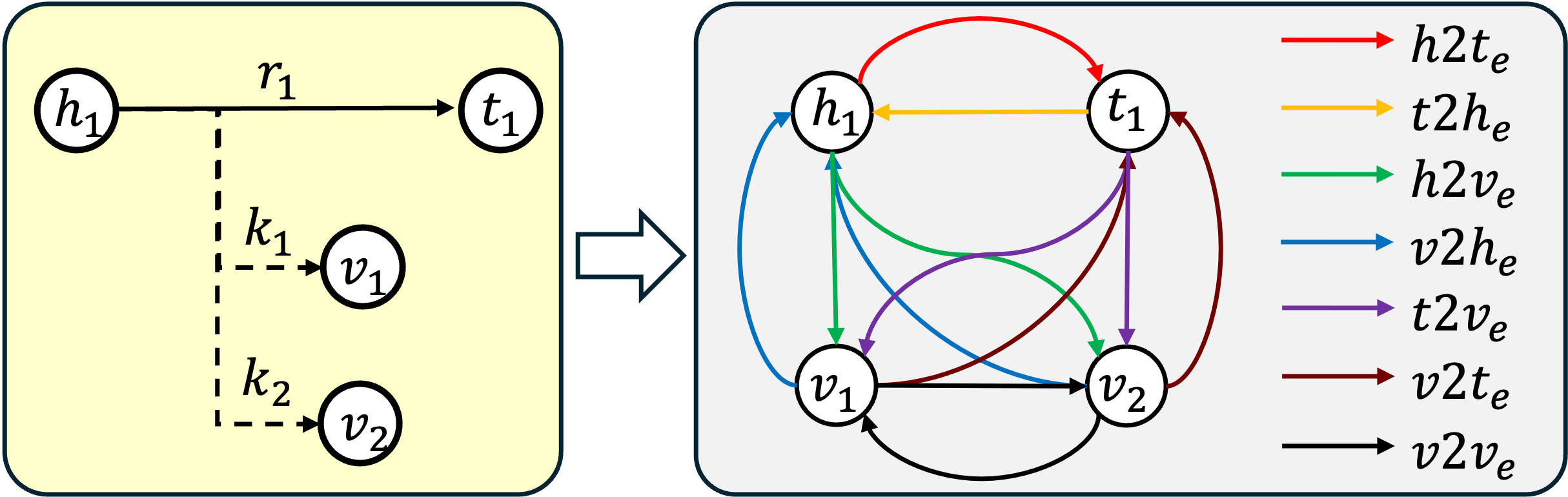}
    \vspace{-1em}
    \caption{Entity foundation graph with different types of fundamental intra-fact interactions.}
    \vspace{-1em}
    \label{entity_foundation}
\end{figure}

\subsection{Relation and Entity Foundation Graphs}
Our relation and entity foundation graphs are designed to capture the fundamental interactions between elements in an HKG, to support the fully-inductive link prediction. To this end, we consider \textit{position-wise interactions} as fundamental interactions, which are agnostic to specific entities and relations.

\textbf{Relation Foundation Graph.}
Inspired by \cite{wei2025inductive, galkin2023towards}, we design a relation foundation graph $\mathcal{G}^r_{fund}$ modeling the fundamental interaction between relations, where two relations are connected according to their semantic positions in the facts. First, we consider $\left \{ h2h_r, h2t_r, t2h_r, t2t_r \right \}$ to depict the relation interactions anchored with primary triplets across different facts. For example, Figure \ref{foundation_relation} shows when two facts share the same head entity, their primary relations $r_1$ and $r_2$ are connected by two reciprocal edges $h2h_r$, indicating their \underline{h}ead-to-\underline{h}ead interactions while the subscript $r$ implies the edge belongs to the \underline{r}elation foundation graph. Second, the qualifiers (represented as k-v pairs) in a hyper-relational fact significantly enrich yet also complicate the relation interactions. Considering the fact that the k-v pairs serve as auxiliary descriptions for the primary triplet, we define $\left \{ r2k_r, k2r_r  \right \}$ to depict the interaction between the primary relation and the keys within a fact, as shown in Figure \ref{foundation_relation}. Meanwhile, we also consider the interaction between keys $k2k_r$ within a fact. Third, we further consider fundamental interactions connecting a key to the relations of other facts via its value entity $v$, i.e., $\left \{ h2v_r, v2h_r, t2v_r, v2t_r, v2v_r  \right \}$, as shown in Figure \ref{foundation_relation}. For example, the primary relation $r_5$ and the key relation $k_2$ are connected by two reciprocal edges $h2v_r$ and $v2h_r$, indicating their \underline{h}ead-to-\underline{v}alue and \underline{v}alue-to-\underline{h}ead interactions, respectively.

\textbf{Entity Foundation Graph.}
We design an entity foundation graph $\mathcal{G}^e_{fund}$ modeling the intra-fact interactions within a hyper-relational fact. Specifically, we first define entity fundamental interactions $\left \{h2t_e, t2h_e \right \}$ for the primary triplet, indicating \underline{h}ead-to-\underline{t}ail and \underline{t}ail-to-\underline{h}ead interactions, respectively, while the subscript $e$ implies the edge belongs to the \underline{e}ntity foundation graph, as shown in Figure \ref{entity_foundation}. Second, we also consider the interactions involving the qualifiers $\left \{h2v_e, v2h_e, t2v_e, v2t_e, v2v_e \right \}$. For example, $h_1$ and $v_1$ are connected by two reciprocal edges $h2v_e$ and $v2h_e$, indicating their \underline{h}ead-to-\underline{v}alue and \underline{v}alue-to-\underline{h}ead interactions, respectively.

In summary, the relation and entity foundation graphs model the fundamental interactions between relations and entities in an HKG, respectively. \textit{These position-wise interactions are agnostic to any specific relations and entities, capturing the structural invariance of HKGs and thus being capable of supporting fully-inductive link prediction.} In our experiments later, we conducted a thorough ablation study to systematically evaluate the usefulness of different types of fundamental interactions, and subsequently select the optimal set of relation fundamental interactions $\mathcal{R}_{fund}^{r} = \{ h2h_r, h2t_r, t2h_r, t2t_r, r2k_r, k2r_r \}$ and entity fundamental interactions $\mathcal{R}_{fund}^{e} = \{h2t_e, t2h_e, h2v_e, v2h_e, t2v_e, v2t_e, v2v_e \}$.

\subsection{THOR}
\label{sec:thor}
\begin{figure*}
    \centering
    \resizebox{\linewidth}{!}{
    \includegraphics[]{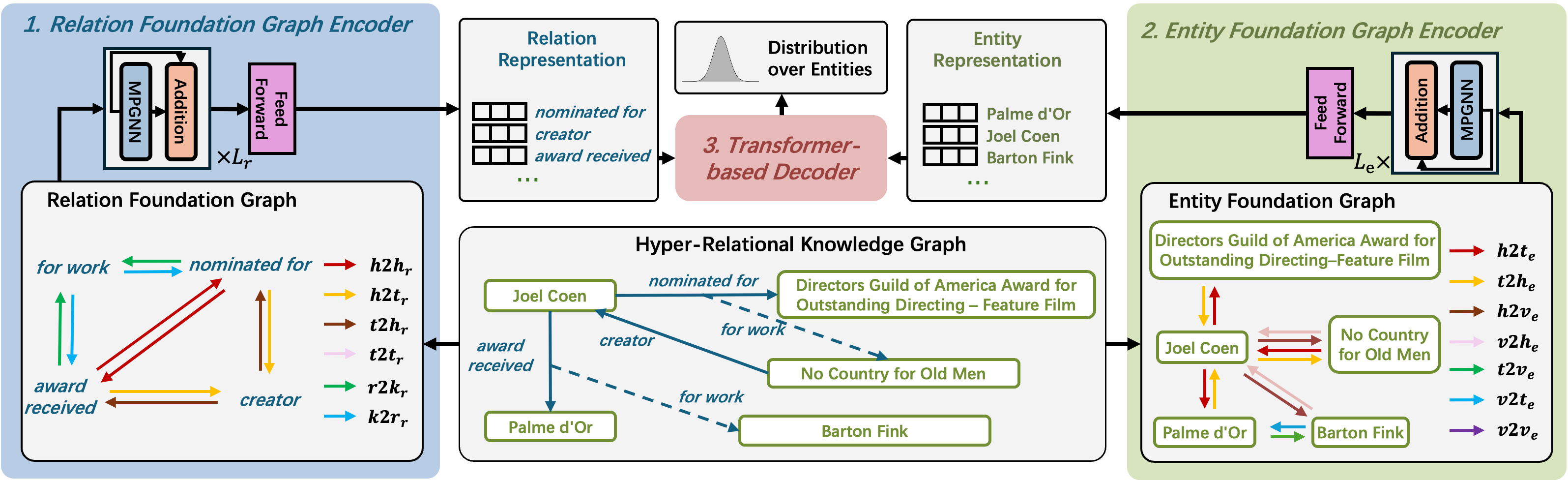}
    }
    \vspace{-2em}
    \caption{Architectural overview of THOR with 1) a relation foundation graph encoder, 2) an entity foundation graph encoder, and 3) a transformer-based decoder.}
    \vspace{-1 em}
    \label{framework}
\end{figure*}

Based on both relation and entity foundation graphs, we design THOR with two parallel graph encoders based on neural Bellman-Ford Networks (NBFNet) \cite{zhu2021neural}, followed by a transformer decoder, which supports efficient masked training and fully-inductive inference. Figure \ref{framework} shows the architectural overview of THOR, which consists of three components. 

\textbf{Relation and Entity Foundation Graph Encoders.}
We adopt NBFNet \cite{zhu2021neural} as graph encoders for both relation and entity foundation graphs. Specifically, different from traditional GNNs that encode node features under the graph automorphism property, where different but automorphic nodes receive the same unconditional feature in message passing (i.e., neighborhood symmetry), NBFNet uses a labeling trick \cite{zhang2021labeling} for relational graphs that provides conditional message passing based on given query nodes. It uses an indicator function to initialize query node features, and then performs message passing to obtain all node representations for the subsequent tasks. In other words, it generates node representations conditioned on a given query, which is more expressive than standard unconditional GNN encoders \cite{zhu2021neural}. We advocate for this labeling trick and adopt NBFNet as encoders in THOR. Formally, given a query hyper-relational fact $q$ with a masked entity, the relation foundation graph encoder first labels all its relations (primary relation and keys) using an indication function with all-ones values, and then performs message passing:
\begin{equation}
    \overrightarrow{h_{r|q}^{0}} = \mbox{INDICATOR} \left ( r, q \right ) = \mathbbm{1}_{r\in q} * \mathbf{1}  ^{d} , r\in \mathcal{G}_{fund}^{r}
\end{equation}
\begin{equation}
\resizebox{\linewidth}{!}{$\overrightarrow{h_{r|q}^{l+1}} = \mbox{UPDATE} \left( \overrightarrow{h_{r|q}^{l}}, \mbox{AGGR} \left( \mbox{MESSAGE} \left( \overrightarrow{h_{w|q}^{l}}, \theta^r \right) | w \in \mathcal{N}_{\mathcal{G}_{fund}^{r}}\left (r \right), \theta^r \in \mathcal{R}_{fund}^{r}   \right)  \right) $}
\end{equation}
where $d$ is the embedding size and $\mathcal{N}_{\mathcal{G}_{fund}^{r}}\left (r \right)$ refers to the neighbors of relation $r$ in the relation foundation graph. The message passing functions follow the design of NBFNet. Note that we use the labeling trick with all-ones values because it is more generalizable in inductive settings than learnable embeddings, which tend to overfit the training graph \cite{huang2023theory}. Similarly, the entity foundation graph encoder first labels all entities (head, tail, and values of qualifiers) of the query fact $q$ using the indication function with all-ones values, and then performs message passing as follows:
\begin{equation}
    \overrightarrow{h_{e|q}^{0}} = \mbox{INDICATOR} \left ( e, q \right ) = \mathbbm{1}_{e\in q} * \mathbf{1}  ^{d} , e\in \mathcal{G}_{fund}^{e}
\end{equation}
\begin{equation}
    \resizebox{\linewidth}{!}{$\overrightarrow{h_{e|q}^{l+1}} = \mbox{UPDATE} \left( \overrightarrow{h_{e|q}^{l}}, \mbox{AGGR} \left( \mbox{MESSAGE} \left( \overrightarrow{h_{w|q}^{l}}, \theta^e \right) | w \in \mathcal{N}_{\mathcal{G}_{fund}^{e}}\left (e \right), \theta^e \in \mathcal{R}_{fund}^{e}   \right)  \right) $}
\end{equation}
Following this design, \textit{our relation and entity foundation graph encoders learn the embeddings of fundamental relation interactions $\theta^r \in \mathcal{R}_{fund}^{r}$ and entity interactions $\theta^e \in \mathcal{R}_{fund}^{e}$, rather than embeddings of specific relations or entities, thus supporting fully-inductive tasks}.

\begin{table*}
\centering
\caption{Dataset Statistics}
\label{statistics}
\vspace{-1em}
\small
\renewcommand{\arraystretch}{0.92}
\begin{tabular}{l|l|rr|rr|rr|rr}\hline
\multicolumn{2}{c}{\multirow{2}{*}{Dataset}} &\multicolumn{2}{|c|}{Train} &\multicolumn{2}{c|}{Inference} &\multicolumn{2}{c|}{Valid} &\multicolumn{2}{c}{Test} \\
\cline{3-10}
\multicolumn{2}{c|}{} & \#Fact & \#Ent/Rel & \#Fact & \#Ent/Rel & \#Fact & \#Ent/Rel & \#Fact & \#Ent/Rel \\
\hline
\multirow{6}{*}{\begin{tabular}[l]{@{}l@{}}Semi-\\inductive \\datasets \end{tabular}} &WD20K100(V1) &7785 &5785/91 &2667 &4286/74 &295 &642/42 &365 &777/42 \\
&WD20K100(V2) &4146 &3226/56 &4274 &5593/53 &539 &975/41 &678 &1215/42 \\
&WD20K66(V1) &9020 &6537/178 &6949 &8367/151 &910 &1515/110 &1114 &1797/109 \\
&WD20K66(V2) &4553 &4286/147 &8922 &9936/119 &1481 &2331/78 &1841 &2714/88 \\
&WD20K33(V1) &23355 &16284/321 &8001 &10277/245 &885 &1208/57 &1095 &1392/59 \\
&WD20K33(V2) &12438 &11335/278 &12822 &14165/228 &1617 &1910/49 &2034 &2200/50 \\
\hline
\multirow{6}{*}{\begin{tabular}[l]{@{}l@{}}Fully-\\inductive\\ datasets \end{tabular}} &WDSPLIT100(V1) &12117 &8927/95 &7676 &9207/303 &1279 &2085/109 &1280 &2103/114 \\
&WDSPLIT100(V2) &7958 &6964/132 &11652 &9551/170 &1942 &2645/76 &1942 &2628/81 \\
&JFFI100(V1) &5325 &3503/147 &7467 &4522/73 &1666 &802/26 &2079 &870/24 \\
&JFFI100(V2) &6330 &2840/126 &9320 &5014/96 &1778 &1146/41 &2217 &1239/41 \\
&JFFI(V1) &21207 &5360/194 &15375 &6495/119 &1708 &872/33 &2137 &966/29 \\
&JFFI(V2) &9012 &3328/156 &18152 &7515/157 &2017 &1464/50 &2523 &1613/51 \\
\hline
\end{tabular}
\vspace{-1em}
\end{table*}

\textbf{Transformer-based Decoder.}
\label{sec:prediction}
After obtaining the conditional entity and relation representations for a given query fact, we adopt a transformer-based decoder for hyper-relational link prediction tasks. Specifically, we adopt an edge-biased self-attention network as a decoder \cite{shaw2018self}, utilizing learnable edge biases to capture the positional interactions of elements (entities and relations) within a hyper-relational fact. Considering the representation of each element (an entity or relation representation learnt above) in a hyper-relational fact as $\overrightarrow{u_i}$, the pair-wise similarity between the elements is computed as follows:
\begin{equation}
    \beta _{ij} =\frac{\left ( W^{Q}\overrightarrow{u_{i} } \right )^{T}\left ( W^{K}\overrightarrow{u_{j} }+\overrightarrow{c_{ij}^{K}}   \right )  }{\sqrt{d} } 
\end{equation}
where $W^{Q}$ and $W^{K}$ are linear transformation matrices for attention query and key, and $c_{ij}^{K}$ indicates learnable edge bias for the key. Following \cite{wang2021link}, we consider five types edge biases, including $(h, r)$, $(t, r)$, $(r, k)$, $(k, v)$, and others not included. After a Softmax layer for normalizing similarity score $\beta_{ij}$, the updated element representation for the next layer can be calculated as follows:
\begin{equation}
    \overrightarrow{u_{i}^{'} } =  \sum_{j=1}^{N}\frac{exp\left ( \beta _{ij}  \right ) }{\sum_{k=1}^{N}exp(\beta _{ik}) } \left ( W^{V}\overrightarrow{u_{j} } + \overrightarrow{c_{ij}^{v}}   \right )  
\end{equation}
where $N$ is the number of elements in a hyper-relational fact, $W^{V}$ is the linear transformation matrix for attention value, and $c_{ij}^{V}$ is the learnable edge bias for value, similar to $c_{ij}^{K}$. After generating the representation of the $[MASK]$ token as $\overrightarrow{x_m}$, the probabilities of the candidates (entities) for the masked entity can be computed as:
\begin{equation}
    \overrightarrow{p} = Softmax\left ( U_0\overrightarrow{x_m} + b_m \right )  
\end{equation}
where $U_0$ is the entity representation matrix generated from the entity foundation graph encoder, and $b_m$ is a learnable bias. 


\textbf{Training and Inference.} 
THOR can be trained in an end-to-end manner by feeding the encoded entity and relation representations to the decoder and minimizing the cross-entropy loss. Notably, our transformer-based decoder can directly score all candidate entities in the entity foundation graphs without the need for expensive negative sampling process such as \cite{zhu2021neural,galkin2023towards} in the training process. 

In the inference stage, given an inference graph and a query hyper-relational fact, THOR first builds the corresponding relation and entity foundation graphs, then encodes both graphs conditioned on the query facts, and finally feeds encoded entity and relation representations to the decoder for predicting the missing entity within the inference graph. The inference stage supports fully-inductive settings by design, because \textit{THOR resolves the link prediction task for a given query on a given KG through the relation and entity foundation graphs built from that KG by learning the fundamental relation and entity interactions, which are agnostic to any specific relations and entities}.


\section{Experiments}
\subsection{Experiment Setup}
\label{sec:setup}

\noindent \textbf{Dataset.}
We evaluate THOR on 12 datasets with different settings. We first use an existing semi-inductive HKG data series WD20K, which consists of 6 datasets of different versions. Due to the lack of fully-inductive HKG datasets in the current literature, we also crawl and process another 6 datasets from two sources with different versions. Table \ref{statistics} summarizes the dataset statistics.
\begin{itemize}[leftmargin=*]
    \item \textbf{WD20K} series \cite{ali2021improving} are semi-inductive HKG datasets collected from Wikidata, with consistent relation vocabulary and disjoint entity vocabularies between training and inference graphs. There are three variants of this dataset with different proportions of hyper-relational facts; WD20K100, WD20K66, and WD20K33 contain 100\%, 66\%, and 33\% hyper-relational facts, respectively. Note that the original WD20K33 has overlapped entities between training and inference graph; we thus filter our own WD20K33 without overlapped entities following the strategy in \cite{ali2021improving} (see Appendix \ref{app_dataset} for details) to avoid data leakage. Following \cite{ali2021improving}, \textit{each dataset variant has two versions, V1 and V2, where V1 has a larger training graph while V2 has a larger inference graph}.
    
    \item \textbf{WDSPLIT100} is our own collected fully-inductive dataset, which retains disjoint entity vocabularies but partially overlapped relation vocabularies between training and inference graphs. Following the inductive KG dataset processing guideline in \cite{shomer2024towards} (originally proposed for triple facts), we first crawl over 1 million hyper-relational facts from \underline{\textbf{W}}iki\underline{\textbf{D}}ata \cite{vrandevcic2014wikidata}, and then \underline{\textbf{SPLIT}} this raw dataset with clustering algorithms on its entity graph to get disjoint entity clusters, and then select the hyper-relational facts whose entities belongs to two different clusters as training and inference graphs (see Appendix \ref{app_dataset} for details). We also adopt a similar two-version design, resulting in WDSPLIT100(V1) and WDSPLIT100(V2). Note that this dataset has 100\% hyper-relational facts, as we crawl only hyper-relational facts by design.
    
    \item \textbf{JFFI} is our own processed dataset filtered from \underline{\textbf{JF}}17K \cite{zhang2018scalable} (originally collected from Freebase) for the hardest \underline{\textbf{F}}ully \underline{\textbf{I}}nductive setting, with both disjoint entity and relation vocabularies between training and inference graphs. Specifically, we use the same process as for the WD20K33 dataset (see Appendix \ref{app_dataset}) to first create training and inference graphs with disjoint entity vocabularies, and then further remove the overlapped relations between training and inference graphs, resulting in a fully-inductive dataset with both disjoint entity and relation vocabularies. We also adopt a similar version design as for other datasets, resulting in JFFI(V1) and JFFI(V2). Moreover, we also consider their variants with 100\% hyper-relational facts, JFFI100(V1) and JFFI100(V2). 
\end{itemize}


\noindent \textbf{Baselines.}
We compare THOR against a sizable collection of baselines in four categories: 1) Transductive hyper-relational link prediction techniques \textbf{StarE} \cite{galkin2020message}, \textbf{GRAN} \cite{wang2021link}, and \textbf{HAHE} \cite{luo2023hahe}; 2) Rule-based inductive techniques \textbf{Neural LP} \cite{yang2017differentiable} and \textbf{DRUM} \cite{sadeghian2019drum}; 3) Semi-inductive techniques \textbf{NBFNet} \cite{zhu2021neural}, \textbf{QBLP} \cite{ali2021improving}, and \textbf{NS-HART} \cite{yin2025inductive}; 4) Fully-inductive techniques \textbf{INGRAM} \cite{lee2023ingram}, \textbf{RMPI} \cite{geng2023relational}, \textbf{ULTRA} \cite{galkin2023towards}, \textbf{MOTIF} \cite{huang2025expressive}, \textbf{KG-ICL} \cite{cui2024prompt}, \textbf{TRIX} \cite{zhang2025trix}, \textbf{MAYPL} \cite{leestructure} and \textbf{MetaNIR} \cite{wei2025inductive}.  Please see Appendix \ref{app_baseline} for more details. Note that among all inductive techniques, only QBLP, NS-HART, MAYPL, and MetaNIR can handle hyper-relational facts; we thus drop qualifiers for all other baselines. Moreover, the original QBLP model assumes pretrained entity and relation embeddings available \cite{reimers2019sentence} (only provided for Wikidata) in the training and inference datasets; this setting contradicts the core idea of inductive settings, where no information except for the inference KG itself is available. Therefore, we also consider the variant of QBLP by removing its pretrained embeddings, denoted as \textbf{QBLP (w/o pretrain)}. In all experiments, we report MRR as the metric.

\begin{table*}
\centering
\caption{Performance comparison with inductive techniques. ``H/T'' refers to the results in head and tail entity prediction; ``ALL'' refers to the results in predicting all entities (including value entity); non-applicable techniques are marked with N/A.}
\label{main_result_inductive}
\small
\renewcommand{\arraystretch}{0.92}
\vspace{-1em}
\begin{tabular}{l|r|rr|rr|rr|rr|rr|rrr}\hline
\multicolumn{2}{c}{\multirow{2}{*}{Method}} &\multicolumn{2}{|c}{WD20K100(V1)} &\multicolumn{2}{|c}{WD20K66(V1)} &\multicolumn{2}{|c}{WD20K33(V1)} &\multicolumn{2}{|c}{WDSPLIT100(V1)} 
&\multicolumn{2}{|c}{JFFI100(V1)} &\multicolumn{2}{|c}{JFFI(V1)} \\
\cline{3-14}
\cline{3-14}
\multicolumn{2}{c|}{} &H/T &ALL &H/T &ALL &H/T &ALL &H/T &ALL &H/T &ALL &H/T &ALL \\
\hline
\multirow{2}{*}{\begin{tabular}[l]{@{}l@{}}Rule-\\based\end{tabular}} &Neural LP &0.2471 &N/A &0.1584 &N/A &0.1638 &N/A &0.0873 &N/A &0.0146 &N/A &0.0170 &N/A \\
&DRUM &0.2834 &N/A &0.1637 &N/A &0.1549 &N/A &0.0759 &N/A &0.0152 &N/A &0.0093 &N/A \\
\hline
\multirow{3}{*}{\begin{tabular}[l]{@{}l@{}}Semi-\\inductive \end{tabular}} &NBFNet &0.2280 &N/A &0.1124 &N/A &0.1050 &N/A &0.0550 &N/A &0.0021 &N/A &0.0029 &N/A \\
&QBLP (w/o pretrain) &0.0052 &N/A &0.0012 &N/A &0.0026 &N/A &0.0034 &N/A &0.0358 &N/A &0.0023 &N/A \\
&QBLP  &0.1028 &N/A &0.0368 &N/A &0.1045 &N/A &0.0045 &N/A &N/A &N/A &N/A &N/A \\
&NS-HART &0.5325 &0.5694 &0.1868 &0.2622 &0.1865 &0.2116 &0.2390 &0.2272 &0.0609 &0.0472 &0.0211 &0.0223 \\
\hline
\multirow{7}{*}{\begin{tabular}[l]{@{}l@{}}Fully-\\inductive \end{tabular}} &INGRAM &0.2420 &N/A &0.0930 &N/A &0.0190 &N/A &0.0680 &N/A &0.0430 &N/A &0.1640 &N/A \\
&RMPI &0.4583 &N/A &0.1791 &N/A &0.1823 &N/A &0.2671 &N/A &0.0595 &N/A &0.0047 &N/A \\
&ULTRA &0.4380 &N/A &0.2359 &N/A &0.1980 &N/A &0.3358 &N/A &0.3144 &N/A &0.3073 &N/A \\
&MOTIF &0.4264&N/A&0.2507&N/A&0.1108&N/A&0.3227&N/A&0.3362&N/A&0.2765&N/A\\
&KG-ICL &0.5284 &N/A &0.2390 &N/A & 0.2644&N/A &0.2786 &N/A &0.0675&N/A &0.0746 &N/A \\
&TRIX &0.2064 &N/A &0.1984 &N/A &0.2205 &N/A &0.3579 &N/A &\textbf{0.3869 }&N/A &0.3122 &N/A \\
&MAYPL & 0.4421& 0.4761&0.1816 &0.2468 &0.2934 &0.3077 & 0.2453&0.2854 &0.1262& 0.1773&0.1331 &0.0967 \\
&MetaNIR &0.3052 & 0.2749& 0.2045& 0.2401&0.2587 &0.2560 &0.1376 &0.1359 &0.2374 & 0.1773&0.2490 &0.1914 \\
\cline{2-14}
&THOR &\textbf{0.6365} &\textbf{0.6557} &\textbf{0.2734} &\textbf{0.3486} &\textbf{0.3066} &\textbf{0.3669} &\textbf{0.4913} &\textbf{0.5125} &0.3687 &\textbf{0.2612} &\textbf{0.3683} &\textbf{0.2638} \\
\hline
\end{tabular}

\begin{tabular}{l|r|rr|rr|rr|rr|rr|rrr}\hline
\multicolumn{2}{c|}{\multirow{2}{*}{Method}} &\multicolumn{2}{c|}{WD20K100(V2)} &\multicolumn{2}{c|}{WD20K66(V2)} &\multicolumn{2}{c|}{WD20K33(V2)} &\multicolumn{2}{c|}{WDSPLIT100(V2)} &\multicolumn{2}{c|}{JFFI100(V2)} &\multicolumn{2}{c}{JFFI(V2)} \\
\cline{3-14}
\cline{3-14}
\multicolumn{2}{c|}{} &H/T &ALL &H/T &ALL &H/T &ALL &H/T &ALL &H/T &ALL &H/T &ALL \\
\hline
\multirow{2}{*}{\begin{tabular}[l]{@{}l@{}}Rule-\\based \end{tabular}} &Neural LP &0.1964 &N/A &0.2053 &N/A &0.1490 &N/A &0.0839 &N/A &0.0394 &N/A &0.0281 &N/A \\
&DRUM &0.1982 &N/A &0.2274 &N/A &0.1338 &N/A &0.0887 &N/A &0.0237 &N/A &0.0249 &N/A \\
\hline
\multirow{3}{*}{\begin{tabular}[l]{@{}l@{}}Semi-\\inductive \end{tabular}} &NBFNet &0.1647 &N/A &0.0845 &N/A &0.0707 &N/A &0.0612 &N/A &0.0029 &N/A &0.0081 &N/A \\
&QBLP (w/o pretrain) &0.0024 &N/A &0.0017 &N/A &0.0042 &N/A &0.0023 &N/A &0.0051 &N/A &0.0016 &N/A \\
&QBLP &0.0521 &N/A &0.0178 &N/A &0.0509 &N/A &0.0091 &N/A &N/A &N/A &N/A &N/A \\
&NS-HART &0.3628 &0.4185 &0.1454 &0.2270 &0.1328 &0.1613 &0.1885 &0.2184 &0.0343 &0.0386 &0.0129 &0.0165 \\
\hline
\multirow{7}{*}{\begin{tabular}[l]{@{}l@{}}Fully-\\inductive \end{tabular}} &INGRAM &0.0760 &N/A &0.0130 &N/A &0.0040 &N/A &0.0360 &N/A &0.0730 &N/A &0.0400 &N/A \\
&RMPI &0.3615 &N/A &0.1831 &N/A &0.1571 &N/A & 0.2654&N/A & 0.0702&N/A &0.0881 &N/A \\
&ULTRA &0.3769 &N/A &0.1945 &N/A &0.2042 &N/A &0.2922 &N/A &0.2167 &N/A &0.2105 &N/A \\
&MOTIF &0.1843&N/A&0.1939&N/A&0.1788&N/A&0.2658&N/A&0.2481&N/A&0.1792&N/A\\
&KG-ICL &0.4041 &N/A &0.2342 &N/A &0.2012 &N/A &0.3578 &N/A &0.1022&N/A & 0.1098&N/A \\
&TRIX &0.4235 &N/A &0.2154 &N/A &0.1758 &N/A &0.3230 &N/A &0.2090 &N/A &0.2445 &N/A \\
&MAYPL & 0.2482& 0.3445& 0.1625& 0.2388&0.2176 &0.2712 &0.2420 & 0.3049&0.1863&0.1613 &0.1859 & 0.1633\\
&MetaNIR &0.2103 &0.2310 & 0.1360&0.1571 &0.1942&0.2068 &0.1856 & 0.2387&0.2887 &0.2155 &0.2376 & 0.1768\\
\cline{2-14}
&THOR &\textbf{0.5097}
&\textbf{0.5904} &\textbf{0.3014} &\textbf{0.3767} &\textbf{0.2284} &\textbf{0.3083} &\textbf{0.4413} &\textbf{0.5411} &\textbf{0.3069} &\textbf{0.2323} &\textbf{0.2777} &\textbf{0.2309} \\
\hline
\end{tabular}
\vspace{-1em}
\end{table*}

\begin{table*}
\centering
\caption{Performance on cross-domain fully-inductive settings (training $\rightarrow$ inference)}\label{cross_domain_results}
\addtolength{\tabcolsep}{-1.5pt}   
\small
\renewcommand{\arraystretch}{0.92}
\vspace{-1em}
\begin{tabular}{l|rr|rr|rr|rr|rr|rr|rr|rrr}\hline
\multirow{2}{*}{Method} &\multicolumn{2}{c|}{\begin{tabular}[l]{@{}l@{}}WD20K33(V1) \\ $\rightarrow$ JFFI(V1) \end{tabular}} &\multicolumn{2}{c|}{\begin{tabular}[l]{@{}l@{}}WD20K33(V1)\\ $\rightarrow$ JFFI(V2) \end{tabular}} &\multicolumn{2}{c|}{\begin{tabular}[l]{@{}l@{}}WD20K33(V2)\\ $\rightarrow$ JFFI(V1) \end{tabular}} &\multicolumn{2}{c|}{\begin{tabular}[l]{@{}l@{}}WD20K33(V2)\\ $\rightarrow$ JFFI(V2) \end{tabular}} &\multicolumn{2}{c|}{\begin{tabular}[l]{@{}l@{}}JFFI(V1) $\rightarrow$ \\ WD20K33(V1) \end{tabular}} &\multicolumn{2}{c|}{\begin{tabular}[l]{@{}l@{}}JFFI(V1) $\rightarrow$ \\ WD20K33(V2) \end{tabular}} &\multicolumn{2}{c|}{\begin{tabular}[l]{@{}l@{}}JFFI(V2) $\rightarrow$ \\ WD20K33(V1) \end{tabular}} &\multicolumn{2}{c}{\begin{tabular}[l]{@{}l@{}}JFFI(V2) $\rightarrow$ \\ WD20K33(V2) \end{tabular}} \\
\cline{2-17}
&H/T &All &H/T &All &H/T &All &H/T &All &H/T &All &H/T &All &H/T &All &H/T &All \\
\hline
ULTRA &0.3075 &N/A &0.2288 &N/A &0.2891 &N/A &0.2121 &N/A &0.1196 &N/A &0.1350 &N/A &0.0904 &N/A &0.0775 &N/A \\
MOTIF &0.0767 &N/A & 0.0777&N/A & 0.1811&N/A & 0.1903&N/A &0.1533 &N/A &0.1234 &N/A &0.1051 &N/A &0.1273 &N/A \\
KG-ICL &0.0803 &N/A &0.1126 &N/A &0.0864 &N/A &0.1148 &N/A &0.2548 &N/A &0.2098 &N/A &0.2164 &N/A & 0.2026&N/A \\
TRIX &0.2957 &N/A &0.1952 &N/A &0.2784 &N/A &0.2178 &N/A &0.1976 &N/A &0.1556 &N/A &0.1498 &N/A &0.1858 &N/A \\
MAYPL &0.1653 &0.1176 & 0.1407&0.1158 &0.1088&0.0803 &0.0896  & 0.0772 & 0.0884 & 0.1221 & 0.0809 & 0.1352 &0.0590  & 0.0975& 0.0571 & 0.1113\\
MetaNIR &0.0671 & 0.0621& 0.0815&0.0840 &0.0817 &0.0532 &0.0786 & 0.0814& 0.0284& 0.0419&0.0415 &0.0518 &0.0269 &0.0289 &0.0265 &0.0380 \\
\hline
THOR &\textbf{0.3165} &\textbf{0.2245} &\textbf{0.2470} &\textbf{0.1776} &\textbf{0.3265} &\textbf{0.2292} &\textbf{0.2444} &\textbf{0.1676} &\textbf{0.2849} &\textbf{0.3399} &\textbf{0.2186} &\textbf{0.3004} &\textbf{0.2796} &\textbf{0.3305} &\textbf{0.2040} &\textbf{0.2842} \\
\hline
\end{tabular}
\vspace{-0.5em}
\end{table*}

\subsection{Performance Comparison with Baselines}

\textbf{Comparison with Inductive Techniques.}
Table \ref{main_result_inductive} shows the comparison with inductive techniques, where THOR achieves the best performance in most cases, yielding 66.1\%, 55.9\%, and 20.4\% improvement over the best-performing rule-based, semi-inductive, and fully-inductive techniques, respectively. Specifically, we see that rule-based techniques and semi-inductive techniques show low performance on fully-inductive datasets (in particular in the hardest inductive setting on JFFI), because these techniques highly rely on the shared relations as anchors for link prediction tasks, which are not always available in fully-inductive settings (completely unavailable on JFFI). Note that rule-based techniques are semi-inductive by nature because they learn the logical rules on relations. Compared to the fully-inductive baselines, which are all designed for triple facts, THOR shows superior performance by effectively learning from the relation and entity foundation graphs designed exclusively for HKGs. Finally, by comparing the performance across different datasets, we observe that THOR's performance generally increases with a higher portion of hyper-relational facts and with larger training graphs.


\begin{table*}
\centering
\caption{Ablation Study}
\label{main_result_abla}
\addtolength{\tabcolsep}{-1.0pt} 
\small
\renewcommand{\arraystretch}{0.92}
\vspace{-1em}
\begin{tabular}{l|r|rr|rr|rr|rr|rr|rrr}\hline
\multicolumn{2}{c}{\multirow{2}{*}{Method}} &\multicolumn{2}{|c}{WD20K100(V1)} &\multicolumn{2}{|c}{WD20K66(V1)} &\multicolumn{2}{|c}{WD20K33(V1)} &\multicolumn{2}{|c}{WDSPLIT100(V1)} 
&\multicolumn{2}{|c}{JFFI100(V1)} &\multicolumn{2}{|c}{JFFI(V1)} \\
\cline{3-14}
\cline{3-14}
\multicolumn{2}{c|}{} &H/T &ALL &H/T &ALL &H/T &ALL &H/T &ALL &H/T &ALL &H/T &ALL \\
\hline
\multirow{8}{*}{\begin{tabular}[l]{@{}l@{}}Relation\\foundation\\graph\end{tabular}} 
&THOR (noR2K) &0.6222 &0.6462 &0.2659 &0.3247 &0.2915 &0.3545 &0.4820 &0.4979 &0.3292 &0.2310 &0.3609 &0.2566 \\
&THOR (noPrim) &0.6162 &0.6378 &0.2653 &0.3298 &0.2839 &0.3459 &0.4819 &0.5005 &0.3232 &0.2289 & 0.3405 &0.2503 \\
&THOR (addK2K) &0.6267 &0.6508 &0.2725 &0.3433 &0.2912 &0.3573 &0.4901 &0.5067 &0.2818 &0.2027 &0.3580 &0.2573 \\
&THOR (addShareV) &0.5890 &0.6366 &0.2551 &0.3180 &0.2940 &0.3586 &0.4793 &0.4962 &0.2485 &0.1836 &0.3398 & 0.2427\\
&THOR (addAllFI) & 0.6205&0.6485&0.2619&0.3333&0.2831&0.3466&0.4859&0.5006&0.3176&0.2301&0.3478&0.2514 \\
&THOR (add3Path) &0.6059&0.6402&0.2677&0.3391&0.2857&0.3511&0.4805&0.5040&0.2829&0.2047&0.3552&0.2571 \\
&THOR (add3Star) &0.6072&0.6246&0.2575&0.3197&0.3002&0.3618&0.4679&0.4965&0.3408&0.2398&0.3611&0.2630\\
&THOR (addAllMotif) &0.6301&0.6517&0.2600&0.3479&0.2889&0.3560&0.4834&0.5038&0.3559&0.2501&0.3358&0.2383 \\
\hline
\multirow{3}{*}{\begin{tabular}[l]{@{}l@{}}Entity\\foundation\\graph \end{tabular}} 
&THOR (noV2V) &0.6290&0.6421&0.2652&0.3414&0.3061&0.3646&0.4831&0.5060&0.3639&0.2574&0.3600&0.2556 \\
&THOR (noP2V) &0.3218&0.2705&0.1210&0.1456&0.1643&0.1686&0.2283&0.2322&0.2674&0.1828&0.2595&0.1839 \\
&THOR (noV)&0.3462&0.2289&0.1156&0.0938&0.1554&0.1346&0.2364&0.1964&0.1792&0.1263&0.2397&0.1633 \\
\hline
\multirow{1}{*}{\begin{tabular}[l]{@{}l@{}}Structure \end{tabular}} 
&THOR (ULTRA-alike) &0.1721 &0.1463 &0.1517 &0.1365 &0.1189 &0.1152 &0.2289 &0.1840 &0.1422 &0.1139 &0.1014 &0.0850 \\
\hline
&THOR &\textbf{0.6365} &\textbf{0.6557} &\textbf{0.2734} &\textbf{0.3486} &\textbf{0.3066} &\textbf{0.3669} &\textbf{0.4913} &\textbf{0.5125} &\textbf{0.3687} &\textbf{0.2612} &\textbf{0.3683} &\textbf{0.2638} \\
\hline
\end{tabular}

\begin{tabular}{l|r|rr|rr|rr|rr|rr|rrr}\hline
\multicolumn{2}{c|}{\multirow{2}{*}{Method}} &\multicolumn{2}{c|}{WD20K100(V2)} &\multicolumn{2}{c|}{WD20K66(V2)} &\multicolumn{2}{c|}{WD20K33(V2)} &\multicolumn{2}{c|}{WDSPLIT100(V2)} &\multicolumn{2}{c|}{JFFI100(V2)} &\multicolumn{2}{c}{JFFI(V2)} \\
\cline{3-14}
\cline{3-14}
\multicolumn{2}{c|}{} &H/T &ALL &H/T &ALL &H/T &ALL &H/T &ALL &H/T &ALL &H/T &ALL \\
\hline
\multirow{8}{*}{\begin{tabular}[l]{@{}l@{}}Relation\\foundation\\graph \end{tabular}}
&THOR (noR2K) &0.4394 &0.4548 &0.2466 & 0.3324&0.2176 & 0.2770&0.3915 & 0.4605&0.2949 &0.2305 &0.1258 &0.1319 \\
&THOR (noPrim) & 0.4581 & 0.4978& 0.2514&0.3253 &0.2231 & 0.2982& 0.3582&0.4424 &0.2504 &0.1900 &0.1867 & 0.1626\\
&THOR (addK2K) & 0.4648&0.4918 &0.2190 &0.2891 &0.2272 &3019 &0.3619 &0.4250 &0.1943 &0.1695 &0.2396 & 0.1939\\
&THOR (addShareV) &0.3699 & 0.3898&0.2318 &0.3392 & 0.2283&0.2991 &0.3414 &0.3934 &0.2687 &0.2064 &0.2314 & 0.1771\\
&THOR (addAllFI) &0.4697&0.5044&0.2362&0.3134&0.2117&0.2908&0.3931&0.4502&0.2716&0.2233&0.1972&0.1710\\
&THOR(add3Path) &0.4864&0.5197&0.2738&0.3508&0.2139&0.2897&0.3798&0.4165&0.2567&0.1999&0.2530&0.1929 \\
&THOR(add3Star) &0.4893&0.5541&0.2759&0.3670&0.2253&0.2985&0.3565&0.3954&0.2932&0.2190&0.2626&0.2072\\
&THOR(addMotif) &0.4878&0.5635&0.2803&0.3562&0.2273&0.3019&0.3861&0.4522&0.2607&0.2164&0.2125&0.1836\\
\hline
\multirow{3}{*}{\begin{tabular}[l]{@{}l@{}}Entity\\foundation\\graph \end{tabular}} 
&THOR (noV2V) &0.5082&0.5548&0.2965&0.3717&0.2260&0.2987&0.4369&0.5108&0.2984&0.227&0.2637&0.2072 \\
&THOR (noP2V) &0.2561&0.2645&0.1236&0.1698&0.1255&0.1596&0.1702&0.2184&0.1953&0.1420&0.1783&0.1317 \\
&THOR (noV) &0.2457&0.1547&0.1123&0.0852&0.1265&0.1060&0.1879&0.1849&0.1448&0.1001&0.1787&0.1237\\
\hline
\multirow{1}{*}{\begin{tabular}[l]{@{}l@{}}Structure \end{tabular}} 
&THOR (ULTRA-alike) &0.2377 &0.1818 &0.1429 &0.1282 &0.1561 &0.1472 &0.1576 &0.1403 &0.1957 &0.1445 &0.1554 &0.1212\\
\hline
&THOR &\textbf{0.5097}
&\textbf{0.5904} &\textbf{0.3014} &\textbf{0.3767} &\textbf{0.2284} &\textbf{0.3083} &\textbf{0.4413} &\textbf{0.5411} &\textbf{0.3069} &\textbf{0.2323} &\textbf{0.2777} &\textbf{0.2309} \\
\hline
\end{tabular}
\vspace{-1em}
\end{table*}

\noindent \textbf{Comparison with Inductive Techniques in cross-domain inference.} We consider cross-domain fully-inductive settings where the training and inference graphs are from different KG domains. As our datasets come from two different KG domains (Wikidata and Freebase)\footnote{Although facts from different KG sources may overlap as they all boil down to general human knowledge, the specific fact representations are different across KGs without any given alignment. We thus still refer to them as the cross-domain setting.}, we select the largest datasets from the two domains (WD20K33 and JFFI) to perform cross-domain fully-inductive link prediction. Table \ref{cross_domain_results} shows the results on all exhaustive dataset combinations, compared to the top fully-inductive baselines. We see that THOR consistently outperforms the baselines, yielding 32.5\% improvement over the best-performing ones, further showing the superiority of THOR in the cross-domain fully-inductive setting.

\textbf{Comparison with Transductive Techniques.} We also compare THOR against SOTA transductive hyper-relational link prediction techniques, in both inductive and transductive settings for the baselines. The results are in Appendix \ref{app_transductive_results} due to space limitations. First, we find that all transductive baselines show very low performance in inductive settings, as expected, because they learn nothing about unseen entities in the inference graphs from the training graph. Second, in the transductive setting where the baselines are trained on the inference graph and evaluated on the test graph, THOR trained in the inductive setting even surprisingly shows a slightly higher performance than the transductive baselines on average, due to its superior generalizability.

\subsection{Ablation Study}

We conduct a series of ablation studies to systematically validate our key design choices in the aspects of 1) relation foundation graphs, 2) entity foundation graphs, and 3) the structure of parallel foundation graph encoders.

\noindent \textbf{Ablation Study on Relation Foundation Graph.}
To validate our design choice for the relation foundation graph, we design the following variants by removing or expanding our selected fundamental interactions $\mathcal{R}_{fund}^{r} = \left \{ h2h_r, h2t_r, t2h_r, t2t_r, r2k_r, k2r_r \right \}$. \textbf{THOR (noR2K)} removes within-fact relation-key interactions connecting the primary \underline{R}elation and \underline{K}eys within the same fact $\left \{ r2k_r, k2r_r \right \} $. \textbf{THOR (noPrim)} removes cross-fact primary relation interactions connecting the \underline{Prim}ary relations across different facts $\left \{ h2h_r, h2t_r, t2h_r, t2t_r \right \} $. \textbf{THOR (addK2K)} adds within-fact key-key interactions connecting \underline{K}eys in different qualifiers within the same fact $k2k_r$. \textbf{THOR (addShareV)} adds cross-fact relation-key interactions connecting relations and keys via \underline{Share}d \underline{V}alues across different facts $\left \{ h2v_r, v2h_r, t2v_r, v2t_r, v2v_r \right \} $. \textbf{THOR (addAllFI)} uses \underline{All} relation \underline{F}undamental \underline{I}nteractions exhaustively, as shown in Figure \ref{foundation_relation}, which is a combination of THOR (addK2K) and THOR (addShareV).

In addition, out of curiosity, we also consider extending the graph motifs, which have recently been shown to be expressive for building triplet-based KG foundation models \cite{huang2025expressive}, to our HKG. To this end, following the original MOTIF framework \cite{huang2025expressive}, we inject additional fundamental interactions if two relations belong to the same motif. We adopt the following suggested motifs (please see \cite{huang2025expressive} for the definition of these motifs). \textbf{THOR (add3Path)} adds interactions connecting relations by sharing \underline{3-Path} motifs. \textbf{THOR (add3Star)} adds interactions connecting relations by sharing \underline{3-Star} motifs. \textbf{THOR (addAllMotif)} adds interactions connecting relations by sharing either the 3-path and 3-star motif, which is a combination of THOR (add3Path) and THOR (add3Star).

Results are shown in Table \ref{main_result_abla}. On one hand, we observe that THOR outperforms THOR (noR2K) and THOR (noPrim) by 11.0\% and 11.1\%, respectively. On the other hand, THOR (addK2K), THOR (addShareV), and THOR (addAllFI) underperform THOR by 11.5\%, 13.6\%, and 9.9\%, respectively. The lessons learnt here are 1) \textit{the cross-fact primary relation interactions and within-fact relation-key interactions encode the key structural invariance transferable across HKGs}, which can effectively support the inductive link prediction tasks; 2) \textit{the within-fact key-key interactions and cross-fact relation-key (also key-key) interactions are redundant and noisy} for capturing such structural invariance, thus resulting in a performance drop in link prediction tasks. 

In addition, we also see that adding motifs surprisingly yields a performance drop of 9.0\%, 6.3\%, and 6.8\% for THOR (add3Path), THOR (add3Star), and THOR (addAllMotif), respectively. This differs from the positive results on triplet-based KGs as reported in \cite{huang2025expressive}. We believe this is probably due to \textit{the motif for triplet-based KGs mismatching the structural invariance encoded in HKGs}, which deserves further investigation in future work.

\noindent \textbf{Ablation Study on Entity Foundation Graph.}
To validate our design choice for the entity foundation graph, we design the following variants by ablating our selected set of fundamental interactions $\mathcal{R}_{fund}^{e} = \left \{h2t_e, t2h_e, h2v_e, v2h_e, t2v_e, v2t_e, v2v_e \right \}$. Specifically, \textbf{THOR (noV2V)} removes the interaction $v2v_e$ connecting two \underline{V}alues within the same fact. \textbf{THOR (noP2V)} removes the interactions connecting head/tail in \underline{P}rimary triplet to any \underline{V}alues in qualifiers $\left \{h2v_e, v2h_e, t2v_e, v2t_e \right \}$. \textbf{THOR (noV)} removes all interactions involving \underline{V}alues in qualifiers $\left \{h2v_e, v2h_e, t2v_e, v2t_e, v2v_e \right \}$. 

Results are shown in Table \ref{main_result_abla}. THOR outperforms THOR (noV2V), THOR (noP2V), and THOR (noV) by 2.5\%, 47.3\%, and 55.6\%, respectively, which verifies our design choice for exhaustive entity interactions in entity fundamental interactions. First, the significant improvement of THOR over THOR (noP2V) and THOR (noV) reveals the predominant role of entity interactions between primary triplets and qualifiers. Moreover, we observe a slight yet noticeable performance decrease when removing $v2v_e$, both from THOR to THOR (noV2V) by 2.5\% and from THOR (noP2V) to THOR (noV) by 15.5\%, which implies that the entity interactions between qualifiers are also important for inductive link prediction.



\noindent \textbf{Ablation Study on the Structure of Parallel Graph Encoders.}
To verify the architectural design of THOR, we reorganize the three components of THOR following the design of a state-of-the-art inductive technique ULTRA \cite{galkin2023towards}, denoted as \textbf{THOR (ULTRA-alike)}. Specifically, it encodes the relation foundation graph first, and then uses the encoded relation representation as the fundamental interactions in the entity foundation graph encoder; subsequently, its training process resorts to negative samples by randomly corrupting entities in positive hyper-relational facts. Note that THOR (ULTRA-alike) uses the same relation foundation graph and the same encoder as THOR. Results are shown in Table \ref{main_result_abla}. THOR significantly outperforms THOR (ULTRA-alike) by 57.9\%, showing the superiority of the THOR architecture in learning HKGs for inductive link prediction.

\section{Related Work}
\subsection{Transductive link prediction over HKGs}
Traditional link prediction techniques for HKGs mainly focus on the transductive setting, predicting within the training vocabulary. Early methods use an n-ary format to represent hyper-relational facts, splitting the primary relation $r$ into $r_h$ and $r_t$, and representing facts as sets of relation-entity pairs. NaLP \cite{guan2019link} and RAM \cite{liu2021role} model the relatedness among all role-value pairs, leveraging structural information for improved performance. Recent works highlight the drawbacks of the n-ary representation, proposing a hyper-relational representation with an unaltered primary triplet and additional qualifiers \cite{rosso2020beyond}. Recently, given the superiority of transformers \cite{wang2021link}, StarE \cite{galkin2020message}, Hy-Transformer \cite{yu2021improving}, QUAD \cite{shomer2023learning}, and HAHE \cite{luo2023hahe} adopt transformer-based decoders and design various encoders to capture correlations among entities and relations. Although these works show superior performance, they fail to accommodate the dynamics of the modern KGs with newly emerging entities and relations.

\subsection{Inductive link prediction over KGs}

Existing works on inductive link prediction can be classified into two categories based on their settings regarding overlapping vocabularies \cite{galkin2023towards}.

\textbf{Semi-inductive techniques} assume a consistent relation vocabulary but arbitrary (even disjoint) entity vocabularies between training and inference KGs. MorsE \cite{chen2022meta} represents unseen entities by averaging the representation of connected relations. RED-GNN\cite{zhang2022knowledge} encodes shared relational digraphs using dynamic programming and employs query-dependent attention to select relevant edges, followed by message passing where unseen entities are initialized with all-zero vectors. NBFNet \cite{zhu2021neural} designs a unified path-based framework using a generalized Bellman-Ford algorithm. Following NBFNet, several path-based techniques integrate selective message passing inspired by the $A^*$ algorithm \cite{zhu2023net, hart1968formal}, information bottleneck \cite{yu2024inductive}, gated message function \cite{liu2023learning}, and path-length-based cutoff \cite{shomer2023distance}. Other semi-inductive techniques encode unseen entities using relational message passing \cite{sun2024query, shao2024expanding}, local relative position within subgraphs \cite{teru2020inductive, mai2021communicative}, and global anchors \cite{xie2024onef}. While most semi-inductive techniques focus on triple facts, some also consider HKGs. QBLP \cite{ali2021improving} extends semi-inductive techniques to HKGs, integrating relational message passing from StarE \cite{galkin2020message}. NS-HART \cite{yin2025inductive} introduces semantic hypergraph sampling and iterative message passing between hyperedges and entities. Although the semi-inductive techniques address the link prediction with unseen entities, their generalizability is still limited by the assumption of a fixed relation vocabulary.

\textbf{Fully-inductive techniques} assume arbitrary (even disjoint) relation and entity vocabularies between training and inference KGs, making them the most generalized case for link prediction. INGRAM \cite{lee2023ingram} builds a relation foundation graph by connecting relations sharing the same head or tail, updating relation representation via relation-graph-based aggregation. RMPI \cite{geng2023relational} proposes four foundational relations and uses relational GAT \cite{velivckovic2017graph} for entity representation. TARGI \cite{ding2025towards} introduces two extra foundation relations, $pair$ and $loop$, to better describe complex topologies. ULTRA \cite{galkin2023towards} adopts RMPI's foundation relation graph and uses neural Bellman-Ford convolutional layers \cite{zhu2021neural} for latent embeddings. TRIX \cite{zhang2025trix} extends ULTRA by designing iterative graph convolutional layers with different orders. MOTIF \cite{huang2025expressive} also extends the relation foundation graph in ULTRA with k-star motif and k-path motif to capture high-order graph features. \cite{gao2023double} unifies INGRAM and ULTRA with a double permutation equivariant graph representation framework. \cite{zhou2023multi} further extends the double equivariance concept to multi-task double equivariance by representing relations with conflicting predictive patterns. KG-ICL \cite{cui2024prompt} tokenizes the k-hop subgraph of entities with specific local structures, and SCORE \cite{wang2024towards} introduces a semantic unifier for features with diversified origins. In this paper, our THOR is also designed under the double equivariance theorem \cite{zhou2023double}, but focuses on accommodating hyper-relational facts, which are increasingly prominent in modern KGs. We present the theoretical analysis of THOR under the double equivariance theorem in Appendix \ref{app_theory}.

\section{Conclusion}
This paper proposes THOR, an induc\underline{T}ive link prediction technique for \underline{H}yper-relational kn\underline{O}wledge g\underline{R}aphs. We first introduce both relation and entity foundation graphs, modeling their fundamental inter- and intra-fact interactions in HKGs, and then design THOR learning from the two foundation graphs with two parallel NBFNet-based graph encoders followed by a transformer decoder, efficiently supporting fully-inductive inference. Our extensive experiments show that THOR outperforms a sizable collection of baselines, yielding 66.1\%, 55.9\%, and 20.4\% improvement over the best-performing rule-based, semi-inductive, and fully-inductive techniques, respectively, and also reveal the key design factors capturing the structural invariance transferable across HKGs for inductive tasks, which we believe shed light on building HKG foundation models.

In the future, we plan to extend the fundamental interactions to accommodate more complex hyper-relational motifs in HKGs.



\bibliographystyle{ACM-Reference-Format}
\bibliography{sample-base}

\appendix



\section{Different Definitions of Inductive Settings}
\label{app_diff_inductive}
In the current literature, a few works such as QBLP \cite{ali2021improving} and NS-HART \cite{yin2025inductive} have a different definition of semi- and fully-inductive link prediction tasks, where their semi-inductive setting refers to the case of a consistent relation vocabulary and \textit{partially overlapped but not completely disjoint} entity vocabularies, while their fully-inductive setting further assumes a consistent relation vocabulary but \textit{completely disjoint} entity vocabularies. In this paper, we follow the mainstream definition of semi- and fully-inductive link prediction settings \cite{galkin2023towards, zhang2025trix, geng2023relational, lee2023ingram, shomer2024towards} as we presented in Section \ref{sec:preliminary}, and list both QBLP \cite{ali2021improving} and NS-HART \cite{yin2025inductive} as semi-inductive techniques.

\section{Theoretical Analysis}
\label{app_theory}
Our design rationale stems from the double equivariance theorem \cite{zhou2023double} (same for ULTRA \cite{galkin2023towards}, MOTIF \cite{huang2025expressive}, and TRIX \cite{zhang2025trix}), where the knowledge graph foundation model should be equivariant to entity or relation permutation, thus generating the same representation for isomorphic subgraphs. Formally, permutation on entities and relations is defined as:

\textbf{Definition A.1 (Entity and relation permutation)}. Given a hyper-relational KG $\mathcal{G}=\left( E, R, F \right)$ with
$F = \left \{ \left [ \left ( h, r, t  \right ), \left \{ \left ( k_{i}, v_{i} \right )   \right \}_{i=1}^{n} \right ] \mid \right. \\ \left. h, t, v_i \in E, r, k_i \in R   \right \} $, entity or relation permutation is $\phi \in \mathbb{S }_e $ and $\tau \in \mathbb{S }_r $ in symmetric groups $\mathbb{S }_e$ and $\mathbb{S }_r$. The operation $\phi \circ \tau \circ \mathcal{G}$ is defined as $\phi \circ \tau \circ \mathcal{G} = \left(\phi \circ E, \tau \circ R,\phi \circ \tau \circ F \right)$, where $(\phi \circ E)_{\phi \circ i} = E_i$, $(\tau \circ R)_{\tau \circ j} = R_j$ and $\phi \circ \tau \circ F = \left \{ \left [ \left (\phi \circ h,\tau \circ r,\phi \circ t  \right ), \left \{ \left (\tau \circ k_{i},\phi \circ v_{i} \right )   \right \}_{i=1}^{n} \right ] \mid \right. \\ \left. \left [ \left ( h, r, t  \right ), \left \{ \left ( k_{i}, v_{i} \right )   \right \}_{i=1}^{n} \right ] \in F   \right \} $.

As reviewed in \cite{zhou2023double}, a fully inductive KG model should generate double equivariant entity and relation representations, defined as:

\textbf{Definition A.2 (Double equivariant knowledge graph representations)}. Given a hyper-relational KG $\mathcal{G}=\left( E, R, F \right)$ with node representations $\Gamma \left ( \mathcal{G}   \right ) $, $\Gamma \left ( \mathcal{G}   \right )$ is double equivariant knowledge graph representations if for any $\phi \in \mathbb{S }_e $ and $\tau \in \mathbb{S }_r $ in relation and entity symmetric groups, $\Gamma \left ( \phi \circ \tau \circ  \mathcal{G}   \right )  = \phi \circ \tau \circ \Gamma \left ( \mathcal{G}   \right )  $.

Following these theorems, our THOR supports fully-inductive reasoning with double-permutation-equivariant representations by 1) designing the entity and relation foundation graphs, where the two graphs' fundamental interactions do not rely on specific entities and relations, respectively, but only according to their semantic positions, thus being \textit{permutation-equivariant}; 2) using NBFNet (an \textit{equivariant} GNN) as graph encoders; and 3) using a transformer decoder (again, \textit{permutation-equivariant by nature}).



\section{Dataset Processing}
\label{app_dataset}
The detailed data processing steps for the \textbf{WD20K33} and \textbf{JFFI} datasets are as follows. Specifically, different from the WDSPLIT100 dataset that we crawl by ourselves, WD20K33 and JFFI are filtered from the existing datasets WD50K \cite{galkin2020message} and JF17K \cite{zhang2018scalable}, respectively. Subsequently, we adopt the filtering strategy proposed in \cite{ali2021improving}. We first sample a number of seed facts from the whole dataset (including train, valid, and test) and their k-hop neighbors, thus forming the training entity set $E_{train}$ and the inductive entity set $E_{ind} = E/E_{train}$. Then we filter facts whose entities belong to $E_{train}$ and $E_{ind}$ into the training dataset $\mathcal{G}_{train}$ and inductive dataset $\mathcal{G}_{ind}$, respectively. At last we split the inductive dataset into inference dataset $\mathcal{G}_{inf}$, valid dataset $\mathcal{G}_{valid}$ and test dataset $\mathcal{G}_{test}$.

The detailed data crawling and processing steps for \textbf{WDSPLIT100} are as follows. Following the inductive KG dataset processing guideline in \cite{shomer2024towards}, we first extract over 1 million hyper-relational facts from WIKIDATA \cite{vrandevcic2014wikidata}, then split this raw dataset with the Louvain clustering algorithm \cite{blondel2008fast} over the graph built by the primary triplets (neglecting their qualifiers at this moment) to get the disjoint entity sets. Afterward, we select the facts whose entities belong to an arbitrary cluster of those entity clusters to form dataset pieces. Finally, we select relatively larger dataset pieces as $\mathcal{G}_{train}$ and $\mathcal{G}_{ind}$. We also adopt a similar two-version design as for WD20K, resulting in WDSPLIT100(V1) and WDSPLIT100(V2). \textit{Note that this dataset has 100\% hyper-relational facts because we crawl only hyper-relational facts by design.}

\begin{table*}[]\centering
\caption{Comparison with transductive baselines in the inductive setting} \label{transductive_baseline_inductive}
\small
\vspace{-1em}
\begin{tabular}{l|rr|rr|rr|rr|rr|rr}\hline
\multirow{2}{*}{Method} &\multicolumn{2}{c|}{WD20K100(V1)} &\multicolumn{2}{c|}{WD20K66(V1)} &\multicolumn{2}{c|}{WD20K33(V1)} &\multicolumn{2}{c|}{WDSPLIT(V1)} &\multicolumn{2}{c|}{JFFI100(V1)} &\multicolumn{2}{c}{JFFI(V1)} \\
\cline{2-13}
&H/T &All &H/T &All &H/T &All &H/T &All &H/T &All &H/T &All \\
\hline
StarE &0.0523 &N/A &0.0106 &N/A &0.0042 &N/A &0.0053 &N/A &0.0458 &N/A &0.0327 &N/A \\
GRAN &0.0417 &0.0482 &0.0176 &0.0234 &0.0081 &0.0096 &0.0047 &0.0058 &0.0677 &0.0493 &0.0487 &0.0373 \\
HAHE &0.0518 &0.0548 &0.0217 &0.0264 &0.0096 &0.0124 &0.0088 &0.0099 &0.0715 &0.0644 &0.0677 &0.0549 \\
\hline
THOR &\textbf{0.6365} &\textbf{0.6557} &\textbf{0.2734} &\textbf{0.3486} &\textbf{0.3066} &\textbf{0.3669} &\textbf{0.4913} &\textbf{0.5125} &\textbf{0.3687} &\textbf{0.2612} &\textbf{0.3683} &\textbf{0.2638} \\
\hline
\end{tabular}

\begin{tabular}{l|rr|rr|rr|rr|rr|rrr}\hline
\multirow{2}{*}{Method} &\multicolumn{2}{c|}{WD20K100(V2)} &\multicolumn{2}{c|}{WD20K66(V2)} &\multicolumn{2}{c|}{WD20K33(V2)} &\multicolumn{2}{c|}{WDSPLIT(V2)} &\multicolumn{2}{c|}{JFFI100(V2)} &\multicolumn{2}{c}{JFFI(V2)} \\
\cline{2-13}
&H/T &All &H/T &All &H/T &All &H/T &All &H/T &All &H/T &All \\
\hline
StarE &0.0437 &N/A &0.0083 &N/A &0.0007 &N/A &0.0103 &N/A &0.0562 &N/A &0.0289 &N/A \\
GRAN &0.0319 &0.0354 &0.0153 &0.0207 &0.0013 &0.0017 &0.0062 &0.0073 &0.0624 &0.0536 &0.0365 &0.0301 \\
HAHE &0.0436 &0.0497 &0.0233 &0.0319 &0.0058 &0.0105 &0.0053 &0.0084 &0.0852 &0.0697 &0.0438 &0.0362 \\
\hline
THOR &\textbf{0.5097}
&\textbf{0.5904} &\textbf{0.3014} &\textbf{0.3767} &\textbf{0.2284} &\textbf{0.3083} &\textbf{0.4413} &\textbf{0.5411} &\textbf{0.3069} &\textbf{0.2323} &\textbf{0.2777} &\textbf{0.2309} \\
\hline
\end{tabular}
\vspace{-0.5em}
\end{table*}



\begin{table*}[]\centering
\caption{Comparison with transductive baselines in the transductive setting} \label{transductive_baseline_transductive}
\small
\vspace{-1em}
\begin{tabular}{l|rr|rr|rr|rr|rr|rrr}\hline
\multirow{2}{*}{Method} &\multicolumn{2}{c|}{WD20K100(V1)} &\multicolumn{2}{c|}{WD20K66(V1)} &\multicolumn{2}{c|}{WD20K33(V1)} &\multicolumn{2}{c|}{WDSPLIT(V1)} &\multicolumn{2}{c|}{JFFI100(V1)} &\multicolumn{2}{c}{JFFI(V1)} \\
\cline{2-13}
\cline{2-13}
&H/T &ALL &H/T &ALL &H/T &ALL &H/T &ALL &H/T &ALL &H/T &ALL \\
\hline
StarE &0.3606 &N/A &0.3021 &N/A &0.2514 &N/A &0.5226 &N/A &0.3336 &N/A &0.3345 &N/A \\
GRAN &0.3964 &0.4257 &\textbf{0.3155} &0.3374 &0.2062 &0.2347 &0.5159 &0.5371 &0.3279 &0.2761 &0.3027 &\textbf{0.2662} \\
HAHE &0.4238 &0.4562 &0.3081 &0.3205 &0.2217 &0.2508 &\textbf{0.5436} &\textbf{0.5709} &0.3401 &\textbf{0.3017} &0.3256 &0.2598 \\
\hline
THOR &\textbf{0.6365} &\textbf{0.6557} &0.2734 &\textbf{0.3486} &\textbf{0.3066} &\textbf{0.3669} &0.4913 &0.5125&\textbf{0.3687} &0.2612 &\textbf{0.3683} &0.2638 \\
\hline
\end{tabular}

\begin{tabular}{l|rr|rr|rr|rr|rr|rrr}\hline
\multirow{2}{*}{Method} &\multicolumn{2}{c|}{WD20K100(V2)} &\multicolumn{2}{c|}{WD20K66(V2)} &\multicolumn{2}{c|}{WD20K33(V2)} &\multicolumn{2}{c|}{WDSPLIT(V2)} &\multicolumn{2}{c|}{JFFI100(V2)} &\multicolumn{2}{c}{JFFI(V2)} \\
\cline{2-13}
\cline{2-13}
&H/T &ALL &H/T &ALL &H/T &ALL &H/T &ALL &H/T &ALL &H/T &ALL \\
\hline
StarE &0.2290 &N/A &\textbf{0.3367} &N/A &0.2166 &N/A &0.5099 &N/A &0.2933 &N/A &0.2672 &N/A \\
GRAN &0.2036 &0.2537 &0.3182 &0.3410 &0.1874 &0.2113 &0.4936 &0.5183 &0.2957 &0.2384 &0.2389 &0.2006 \\
HAHE &0.2217 &0.2496 &0.3264 &0.3459 &0.2054 &0.2183 &\textbf{0.5211} &0.5398 &\textbf{0.3128} &\textbf{0.2740} &0.2673 &0.2245 \\
\hline
THOR &\textbf{0.5097}
&\textbf{0.5904} &0.3014 &\textbf{0.3767} &\textbf{0.2284} &\textbf{0.3083} &0.4413 &\textbf{0.5411} &0.3069 &0.2323 &\textbf{0.2777} &\textbf{0.2309} \\
\hline
\end{tabular}
\vspace{-0.5em}
\end{table*}

\section{Time Complexity}
THOR's time complexity is $O(a^2ed+vd^2)$ composed of NBFNet-based relation and entity encoders with time complexity $O(ed+vd^2)$ and $O(a^2ed+vd^2)$ as well as transformer decoder with time complexity $O(a^2ed)$, where $a$,$e$,$v$, and $d$ are the average arity of hyper-relational facts, the number of facts, the number of entities, and the embedding dimension, respectively. We ignore the relation foundation graph, which is much smaller than the entity foundation graph and also processed in parallel. Compared to ULTRA's complexity $O(e*d+v*d^2)$, THOR is reasonably more complex as it also learns from qualifiers. As an example, on JFFIv1, THOR requires 2~3x more computation than ULTRA(104s/epoch v.s. 48s/epoch in training, 18s v.s. 6s in inference), but yields 19.9\% performance improvement.

\section{Baselines}
\label{app_baseline}
We present the details on our baselines below.

\textbf{Transductive hyper-relational link prediction techniques.} 
\textbf{StarE} \cite{galkin2020message} represents a hyper-relational fact as a directed heterogeneous graph on triplets and designs an attributed message passing GNN where qualifiers are viewed as edge attributes and merged into the primary relation. 
\textbf{GRAN} \cite{wang2021link} is a graph-based approach to link prediction on n-ary relational facts, which effectively captures rich associations within facts by representing them as heterogeneous graphs and utilizing edge-biased fully-connected attention to model local dependencies. 
\textbf{HAHE} \cite{luo2023hahe} maintains the main structure of GRAN while introducing bidirectional hyper-edge message passing to capture the global structural information.

\textbf{Rule-based link prediction techniques}.
\textbf{Neural LP} \cite{yang2017differentiable} combines the parameter and structure learning of first-order logical rules in an end-to-end differentiable model, where inference tasks can be compiled into sequences of differentiable operations. A neural controller system is then developed to compose these operations for link prediction.
\textbf{DRUM} \cite{sadeghian2019drum} is motivated by making a connection between learning confidence scores for each rule and low-rank tensor approximation. It uses bidirectional RNNs to share useful information across the tasks of learning rules for different relations.
Note that in rule-based methods, the filter operation eliminates all unseen entities prior to evaluation, which artificially inflates accuracy in an inductive setting. To ensure a fair comparison, we modify this operation to retain the ground-truth entities present in the test set for ranking only.

\textbf{Semi-inductive link prediction techniques}.
\textbf{NBFNet} \cite{zhu2021neural} generalizes the Bellman-Ford algorithm and proposes a path-based message passing mechanism with labeling trick \cite{zhang2021labeling}, which can be applied to both transductive and inductive settings.
\textbf{QBLP} \cite{ali2021improving} adopts a StarE encoder with a transformer decoder and can reason over an inference hyper-relational graph.
\textbf{NS-HART} \cite{yin2025inductive}. NS-HART defines an n-ary semantic hypergraph with a specifically designed k-hop neighborhood for HKG, and introduces a subgraph aggregator network to model complicated interactions in the subgraphs.

\textbf{Fully-inductive link prediction techniques}.
\textbf{INGRAM} \cite{lee2023ingram} generates a latent representation with relation graphs whose adjacent matrix $E_h$ is defined as $E_h[i, j]$ for times of the $i$th entity connected with the $j$th relation as head and similar for $E_t$. GNN is conducted over relation graphs at both entity-level and relation-level to generate the representation of entities and relations for further link prediction.
\textbf{RMPI} \cite{geng2023relational} defines a relation foundation graph with foundation interaction $h2h_r$, $h2t_r$, $t2h_r$, and $t2t_r$ similar to those in our relation foundation graph. Relational graph convolution is conducted over the relation foundation graph and the original KG for the latent representation of relations and entities.
\textbf{ULTRA} \cite{galkin2023towards} inherits the structure design of the relation foundation graph from RMPI, replaces the convolution layers with NBFNet for path-based convolution, and enhances the model with labeling trick \cite{zhang2021labeling}.
\textbf{MOTIF} \cite{huang2025expressive} extends relation foundation graph of ULTRA with k-star and k-path motif, thus capturing high-order features of KGs. 
\textbf{KG-ICL} \cite{cui2024prompt} extracts prompt graphs for query relations and designs a unified graph tokenizer representing the entities and relations into pre-defined tokens, followed by two message passing neural networks for prompt encoding and KG reasoning.
\textbf{TRIX} \cite{zhang2025trix} is a fully-inductive technique that improves expressivity and relation prediction efficiency by constructing a relation adjacency matrix incorporating entity information and employing an iterative embedding update mechanism to refine representations of both entities and relations.
\textbf{MAYPL} \cite{leestructure} initializes all entities and relations with two learnable entity and relation tokens with neighborhood aggregation to generate entity and relation representations, followed by hyper-edge level message passing for reasoning.
\textbf{MetaNIR} \cite{wei2025inductive} builds complicated inter- and intra-fact relation foundation graph for relational GNN to generate entity and relation representations, followed with GRAN \cite{wang2021link} decoder.

\section{Performance Comparison with Transductive Baselines}
\label{app_transductive_results}
We compare our THOR with transductive baselines StarE \cite{galkin2020message}, GRAN \cite{wang2021link}, and HAHE \cite{luo2023hahe} in both inductive and transductive settings. 

In the inductive setting, unseen entities and relations are not learnt in the training stage and thus have randomly initialized embeddings in the inference stage. The results are shown in Table \ref{transductive_baseline_inductive}, where the transductive baselines yield very low performance as expected. Those transductive techniques highly rely on the well-trained embeddings on the fixed entity and relation vocabulary while learn nothing about unseen entities and relations in the training procedure, thus losing the generalizability to unseen entities or relations in the inductive setting. 

In the transductive setting, the baselines are trained on the inference graph $\mathcal{G}_{inf}$, then validated and evaluated on the valid and test graphs\footnote{Note that these three graphs have consistent relation and entity vocabularies.}, respectively, without using the training graph at all. Table \ref{transductive_baseline_transductive} shows the results compared to our THOR (the results of THOR are still in the fully-inductive setting). \textit{We observe that THOR even surprisingly shows a slightly higher performance (with an improvement of 17.1\% on average) than the transductive baselines, yielding the best results in the majority of the cases, due to its superior generalizability.}

\section{Limitations}
\label{sec:limitation}
Although the proposed THOR demonstrates superior performance in fully-inductive link prediction over HKGs, it still exhibits certain limitations. First, THOR may experience performance degradation on transductive datasets, in particular on domain-specific KGs. This issue could be mitigated by exploring hybrid approaches that integrate both inductive and transductive link prediction paradigms. Second, the stability of THOR warrants further enhancement, as inductive models are generally prone to overfitting, a trend observed in our experiments across both semi- and fully-inductive baselines.

\end{document}